%% file: LaTeX/tmi.tex
\documentclass[journal,twoside]{IEEEtran}

\usepackage[dvipsnames,svgnames,x11names,table]{xcolor}

\usepackage{cite}
\usepackage{amsmath,amssymb,amsfonts}
\usepackage{algorithmic}
\usepackage{graphicx}
\usepackage{textcomp}
\usepackage{multirow}
\usepackage{booktabs} 
\usepackage{subcaption}
\usepackage{wrapfig}

\begin{document}
\title{Patient-Conditioned Adaptive Offsets for Reliable Diagnosis across Subgroups}
\author{Gelei Xu, Yuying Duan, Jun Xia, Ruining Deng, Wei Jin, and Yiyu Shi, \textit{Senior Member, IEEE}
\thanks{G. Xu, J. Xia, Y. Shi are with the Department of Computer Science and Engineering, University of Notre Dame, Notre Dame, IN, 46556, USA. Email: \{gxu4, jxia4, yshi4\}@nd.edu}
\thanks{Y. Duan is with the Department of Electrical Engineering, University of Notre Dame, Notre Dame, IN, 46556, USA. Email: yduan2@nd.edu}
\thanks{R. Deng is with Weill Cornell Medicine, New York, NY, 10044, USA. Email: rud4004@med.cornell.edu}
\thanks{W. Jin is with the Department of Computer Science, Emory University, Atlanta, GA 30322, USA. Email: wei.jin@emory.edu}
}

\maketitle

\begin{abstract}
AI models for medical diagnosis often exhibit uneven performance across patient populations due to heterogeneity in disease prevalence, imaging appearance, and clinical risk profiles. Existing algorithmic fairness approaches typically seek to reduce such disparities by suppressing sensitive attributes. However, in medical settings these attributes often carry essential diagnostic information, and removing them can degrade accuracy and reliability, particularly in high-stakes applications. In contrast, clinical decision making explicitly incorporates patient context when interpreting diagnostic evidence, suggesting a different design direction for subgroup-aware models.
In this paper, we introduce HyperAdapt, a patient-conditioned adaptation framework that improves subgroup reliability while maintaining a shared diagnostic model. Clinically relevant attributes such as age and sex are encoded into a compact embedding and used to condition a hypernetwork-style module, which generates small residual modulation parameters for selected layers of a shared backbone. This design preserves the general medical knowledge learned by the backbone while enabling targeted adjustments that reflect patient-specific variability. To ensure efficiency and robustness, adaptations are constrained through low-rank and bottlenecked parameterizations, limiting both model complexity and computational overhead.
Experiments across multiple public medical imaging benchmarks demonstrate that the proposed approach consistently improves subgroup-level performance without sacrificing overall accuracy. On the PAD-UFES-20 dataset, our method outperforms the strongest competing baseline by 4.1\% in recall and 4.4\% in F1 score, with larger gains observed for underrepresented patient populations.

\end{abstract}

\begin{figure}[hbpt]
  \centering
  \includegraphics[width=0.49\textwidth]{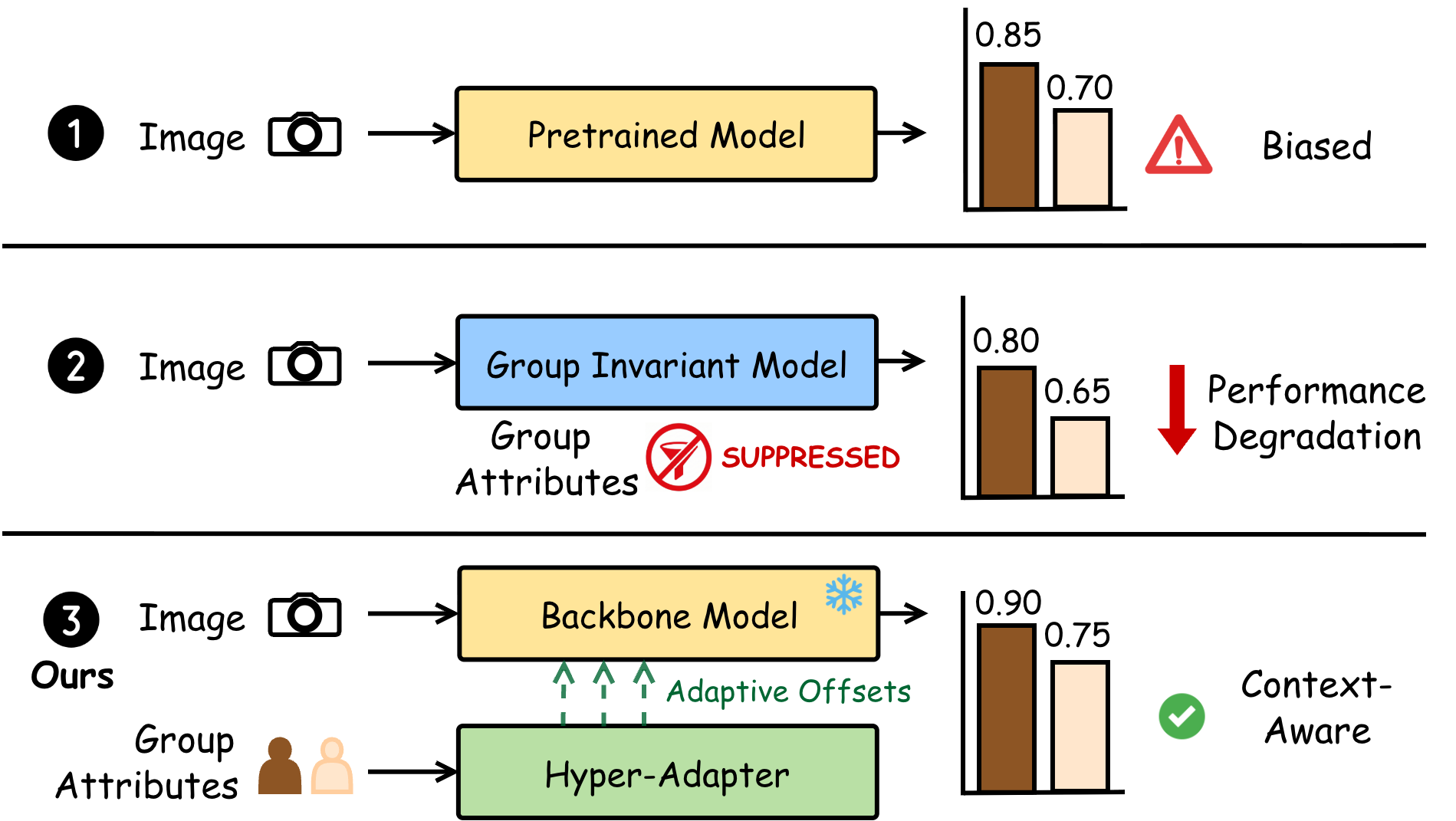}
  \caption{Comparison of paradigms for integrating patient attributes. (1) Standard training leads to biased performance; (2) Group-blind suppression degrades overall accuracy; (3) Our HyperAdapt enables context-aware reliability.}
  \vspace{-14pt}
  \label{fig:abstract}
\end{figure}


\input{tex/introduction}
\input{tex/related_works}
\input{tex/method}
\input{tex/experiments}
\input{tex/conclusion}

\bibliographystyle{IEEEtran}
\bibliography{ref.bib}

\end{document}

%% file: tex/introduction.tex
Medical-AI systems often report high overall accuracy on curated benchmarks, yet this metric alone is insufficient to characterize real-world reliability. In practice, patient groups differ substantially in disease prevalence, imaging appearance, and underlying risk profiles, resulting in systematic variation in the visual and contextual cues required for accurate diagnosis. Standard training pipelines typically optimize a single model over aggregated data, implicitly assuming a shared underlying distribution across patient groups. As a result, population-specific patterns are diluted during training, leading to systematic performance disparities across subgroups. Models tend to perform well on majority or well-represented populations, while predictions for underrepresented groups are often less accurate or less reliable~\cite{seyyed2021underdiagnosis, daneshjou2022disparities}. Such group-dependent errors risk exacerbating existing healthcare disparities and eroding patient trust in AI-assisted diagnosis. Consequently, a central challenge for clinical deployment is to ensure that diagnostic models achieve high average accuracy while maintaining consistently acceptable performance for each subgroup.
%For example, in dermatology, individuals with lighter skin have lower melanin levels, making them more susceptible to melanoma~\cite{caini2009meta}. 

A common response to subgroup performance disparities has been work on algorithmic fairness, which aims to reduce group-dependent errors by suppressing or removing the influence of subgroup attributes such as age and sex~\cite{xu2022algorithmic, xu2024addressing}. In application domains such as recidivism risk assessment~\cite{flores2016false} or income prediction~\cite{asuncion2007uci}, these attributes often act as spurious correlates, and limiting their influence can effectively mitigate bias and improve statistical fairness metrics. 
In medical diagnosis, however, this assumption does not generally hold. Many subgroup attributes are clinically meaningful cues that shape disease presentation, progression, and risk stratification~\cite{caini2009meta}. For example, in dermatology, skin type affects lesion appearance and alters the visual cues used for diagnosis~\cite{narayanan2010ultraviolet}. Suppressing such attributes can therefore remove information essential for diagnostic reasoning, potentially degrading performance across patient groups, including those already underserved. 
Rather than enforcing invariance to patient attributes, an alternative perspective is to view diagnosis as a context-conditioned prediction problem, in which patient attributes inform disease priors and modulate the relevance of specific visual cues~\cite{sabuncu2025ethical, puyol2021fairness, zhang2022improving}. This perspective aligns with clinical practice, where clinicians routinely adjust their interpretation of imaging findings based on patient context.

Existing efforts to incorporate patient attributes into diagnostic models mainly fall into two categories. 
One line of work introduces subgroup-specific models or branches to capture group-dependent patterns~\cite{xu2025incorporating, puyol2021fairness}. While effective at modeling in-group variations, such approaches tend to fragment medical knowledge across separate components and struggle to balance specialization with generalization in realistic clinical settings.
A second line of work incorporates patient attributes through multimodal fusion, combining clinical metadata with imaging representations within a shared backbone~\cite{baltruvsaitis2018multimodal, kline2022multimodal}. Although fusion-based methods preserve parameter sharing, they often treat patient attributes as auxiliary features rather than context that actively shapes diagnostic reasoning, limiting their ability to modulate which visual cues are emphasized for different patients.
Together, these limitations highlight the challenge of incorporating patient attributes in a way that supports context-sensitive diagnosis while preserving a unified representation of medical knowledge.

% While effective in limited settings, this design choice introduces a tension between specialization and generalization. Many clinically meaningful subgroups contain few samples, making specialized models prone to overfitting and unstable calibration. In addition, continuous attributes such as age blur subgroup boundaries and limit the applicability of discrete subgroup definitions. Moreover, interactions among patient attributes can rapidly lead to a combinatorial growth of subgroups that cannot feasibly be supported by separate models or branches.
% These challenges point to the need for a single backbone model that can adapt flexibly and efficiently to patient attributes, while preserving shared medical knowledge and avoiding fragmentation of data into many small groups.

To address the aforementioned challenges, we propose HyperAdapt, a framework that shifts the paradigm from group-invariance to context-conditioning (Figure~\ref{fig:abstract}), enabling patient-conditioned modulation on top of a pretrained backbone. HyperAdapt encodes patient subgroups into a compact embedding, which is then used to generate small residual offsets for selected layers of the backbone. These offsets modulate intermediate computations rather than replacing them, allowing the model to adjust its behavior based on patient context while preserving shared medical representations learned during pretraining.
Directly mapping low-dimensional attribute vectors to full-dimensional parameter offsets would be parameter-intensive and prone to overfitting, particularly under limited subgroup data. To mitigate this issue, we explore several strategies to constrain model complexity, including low-rank factorization, parameter sharing across adaptation offsets, and weight sharing within the hyper-adapter. Together, these design choices enable efficient conditioning on patient attributes without fragmenting the model or substantially increasing its parameter count.
Extensive experiments demonstrate the effectiveness of the proposed approach. On the PAD-UFES-20 dataset, our method improves recall by 4.1\% and F1 score by 4.4\% over the strongest competing baseline, with larger gains observed for underrepresented patient groups. These results show that explicitly modeling clinically meaningful heterogeneity can improve subgroup-level reliability while maintaining strong overall performance. More broadly, the proposed framework offers a scalable and efficient pathway for developing patient-conditioned medical AI systems that better align with the structure of clinical decision making.

%% file: tex/related_works.tex
\section{Related Works}

\subsection{Subgroup Reliability in Medical AI: Fairness, Metadata Modeling, and Personalization}

Medical AI systems frequently exhibit uneven diagnostic performance across patient subgroups, raising growing concerns about reliability, safety, and equity in clinical practice. Substantial prior works have sought to understand and address subgroup-level performance disparities from different methodological perspectives.
This section reviews three complementary lines of research relevant to subgroup performance in medical AI. The first line focuses on algorithmic fairness methods that aim to reduce performance gaps between groups by enforcing parity or invariance across subgroup attributes. The second line targets explicit improvement of subgroup outcomes through subgroup-wise optimization objectives or personalized modeling strategies that tailor predictions to specific patient segments. The third line treats patient attributes as auxiliary signals or metadata and incorporates them through multimodal or metadata-conditioned modeling frameworks.

%Our work is closely related to all three directions. In the experimental evaluation, we select representative methods from each category as baselines, enabling a systematic comparison between alternative approaches to modeling subgroup-specific variability and the proposed patient-conditioned adaptation framework.

\subsubsection{Subgroup Fairness Through Gap Reduction}
Fairness research traditionally frames subgroup performance as a disparity-minimization problem, aiming to optimize metrics such as Equalized Opportunity and Equalized Odds~\cite{hardt2016equality}. Existing approaches operate at different stages of the learning pipeline—including modifying training data distributions, constraining model representations through optimization objectives, or adjusting decision thresholds at inference time—using techniques such as reweighting, adversarial learning, and group-specific post-processing~\cite{kamiran2012data, zhang2018mitigating, xu2025fair}.
Despite their mathematical appeal, gap-reduction methods face intrinsic limitations in clinical settings. First, fairness constraints often lie on the Pareto frontier: improving performance for one group may degrade another~\cite{dehdashtian2024utility}, and some techniques reduce accuracy for all groups~\cite{wu2022fairprune, duan2025cost}. Second, many techniques operate by suppressing or downweighting sensitive attributes, yet in medicine such attributes frequently carry clinically meaningful information. For example, skin type reflects differences in melanin content and UV susceptibility~\cite{caini2009meta}, and disease prevalence varies systematically across demographic groups~\cite{gordon2013skin}. Removing these signals conflicts with clinical reasoning and can compromise diagnostic utility.

\subsubsection{Subgroup-Maximization and Personalized Modeling}
Recognizing these limitations, recent work reframes fairness as maximizing performance within each subgroup rather than minimizing disparities across them. Early studies trained independent models per demographic group~\cite{puyol2021fairness}. Stratified ERM~\cite{zhang2022improving} and decoupled classifiers~\cite{dwork2018decoupled} extend this paradigm by training separate risk minimizers or classifier heads for each subgroup. MEDFAIR~\cite{zong2022medfair} formalizes this direction as domain-specific modeling aimed at capturing subgroup-specific visual patterns. Similar ideas appear in personalized federated learning~\cite{luo2022adapt, tan2022towards} and multi-task learning frameworks~\cite{evgeniou2004regularized}, which balance global knowledge sharing with local specialization.
While subgroup-maximization improves in-group performance, these approaches scale poorly in clinical environments. Explicit subgroup models fragment the parameter space, require substantial in-group data, and cannot generalize to continuous or compositional patient attributes (e.g., age × sex × skin type). Because each subgroup is treated as a static domain, such methods also lack the flexibility to adapt to unseen or sparsely represented combinations of patient factors.

\subsubsection{Metadata-Level Modeling via Multimodal Fusion}
Another family of approaches incorporates patient attributes directly into the model through multimodal fusion. Clinical metadata such as age, sex, lesion location, or other tabular features are combined with image representations through early fusion, late fusion, or joint fusion architectures~\cite{baltruvsaitis2018multimodal, kline2022multimodal}. More advanced fusion techniques, including FiLM-based modulation~\cite{perez2018film} and medical variants such as DAFT~\cite{wolf2022daft}, condition intermediate visual features on tabular inputs. Fusion methods incorporate patient attributes by treating them as a parallel feature stream that is merged with imaging representations. In this formulation, clinical metadata enter the model alongside visual features, and the backbone remains shared across all patients. As a result, the computation path is identical for every subgroup with metadata influencing the learned representation.

Unlike prior approaches to subgroup reliability, our method treats patient attributes as a conditioning context that dynamically shapes how diagnostic evidence is interpreted, which is similar to how clinicians interpret subgroup signals. Our experimental results demonstrate consistent performance improvements over baseline methods, with the largest gains observed for underrepresented patient groups.
%Fusion methods effectively leverage rich metadata and are widely used in multimodal medical systems. However, their influence remains at the feature level: the backbone parameters are shared across all patients, and the model performs the same computation regardless of subgroup. From a clinical perspective, this structure conflicts with diagnostic reasoning, where patient attributes do not serve as parallel evidence streams but instead modulate how visual findings are interpreted~\cite{kassirer1989diagnostic}. Fusion enriches the input representation but does not support patient-conditioned changes in the diagnostic function, limiting its ability to capture subgroup-specific visual patterns or continuous clinical variability.

%Taken together, existing approaches provide only partial solutions: gap-reduction methods promote statistical parity but risk removing clinically informative signals; subgroup-specific models tailor predictions but do not scale; and multimodal fusion incorporates metadata but cannot adapt inference behavior to individual patients. These limitations highlight the need for models in which patient attributes modulate the underlying diagnostic computation rather than merely augmenting the input. A patient-conditioned adaptation mechanism offers a principled path toward reliable, context-sensitive diagnostic performance across diverse populations.

\subsection{Dynamic Neural Networks}
Most neural networks operate in a static manner in which a fixed set of parameters is applied to all inputs without considering instance-level variations. Dynamic neural networks relax this constraint by adapting portions of the computation based on the input. This enables conditional representations that can better capture heterogeneous data distributions~\cite{jia2016dynamic}. Such adaptability is particularly valuable in medical AI, where diagnostic appearance and underlying disease priors often vary across demographic and clinical subgroups.

\subsubsection{Routing-Based Dynamic Networks}
A prominent class of dynamic architectures relies on routing mechanisms to activate different computational paths for different inputs. Mixture-of-Experts models and related routing networks employ gating functions that select a subset of experts conditioned on the input, enabling conditional computation and high model capacity~\cite{shazeer2017outrageously, zhang2023robust}. These approaches include both soft and hard routing variants and other forms of input-dependent path selection. Routing decisions are typically made at the module or expert level and are discrete in nature, which aligns well with coarse-grained specialization. At the same time, this formulation is less naturally suited to representing continuous attributes whose effects vary smoothly (e.g., age)~\cite{xu2025incorporating}. In addition, discrete routing implicitly partitions the training data across experts, meaning each path is updated with only a subset of samples. In settings where subgroup data are limited, such data fragmentation can influence optimization dynamics and calibration behavior.

\subsubsection{Feature Modulation Networks}
Another line of dynamic architectures modulates intermediate features or convolutional filters based on input-dependent signals. Representative examples include dynamic convolution~\cite{jia2016dynamic}, feature-wise linear modulation~\cite{perez2018film}, channel reweighting mechanisms~\cite{hu2018squeeze}, and context-conditioned affine transformations such as conditional batch normalization~\cite{de2017modulating}. These methods adapt scaling, shifting, or filtering operations to capture input-specific characteristics while keeping the underlying network weights fixed. This design offers computational efficiency and has proven effective in a wide range of vision tasks. At the same time, because the modulation operates at the feature-activation level rather than through reparameterization of model weights, the form of functional variation they can induce is restricted.

\subsubsection{Hypernetworks for Conditional Parameter Generation}
Hypernetworks generate parameters for a primary network using a secondary network conditioned on a low-dimensional context vector~\cite{ha2016hypernetworks}. This formulation allows the main model to instantiate different parameterizations for different inputs. Hypernetwork-based conditioning has demonstrated strong performance in tasks such as noise-dependent image restoration~\cite{aharon2023hypernetwork}, single-image 3D reconstruction~\cite{littwin2019deep}, continual learning~\cite{von2019continual}, and personalized federated optimization~\cite{shamsian2021personalized}.  In the medical domain, hypernetwork-based approaches have been explored for tabular personalization through models such as HyperTab~\cite{wydmanski2023hypertab}, which generate ensembles of subnetworks specialized for different low-dimensional views of the data, illustrating the potential of conditional parameter generation for individualized modeling. %Compared with routing-based architectures, hypernetworks allow fine-grained conditioning without constructing group-specific modules. Compared with feature-level modulation, hypernetworks directly adjust model parameters, enabling a broader range of functional variation.

However, directly applying hypernetwork formulations to pretrained medical imaging backbones raises several challenges that limit their practicality. First, the gap between low-dimensional patient metadata and high-dimensional convolutional weights makes direct parameter generation computationally demanding and susceptible to overfitting, often degrading performance in practice. Second, standard hypernetwork architectures typically require the primary network’s parameters to be generated or significantly modified during training, complicating the use of pretrained initialization—an essential component of modern medical imaging pipelines~\cite{kim2022transfer}.
The present work focuses on addressing these challenges by developing a patient-conditioned adaptation mechanism that preserves pretrained backbones while enabling efficient, data-conscious parameter modulation. 
 
 %To address these limitations, we draw on structured parameterization techniques from parameter-efficient adaptation~\cite{hu2022lora, houlsby2019parameter} and use them to regularize hypernetwork outputs. Constraining updates to low-rank or channel-wise components enables the hypernetwork to operate within a compact and stable parameter space that aligns with clinically meaningful metadata while avoiding the instability of generating full-weight tensors. In addition, generating offset weights rather than complete parameters~\cite{guo2021parameter} allows the framework to leverage pretrained backbones and fine-tune them in a controlled manner. This design preserves the expressive capacity of conditional parameter generation while supporting robust and scalable adaptation within medical imaging pipelines.

%% file: tex/method.tex
\section{Method}

\begin{figure*}[t]
  \centering
  \includegraphics[width=0.85\textwidth]{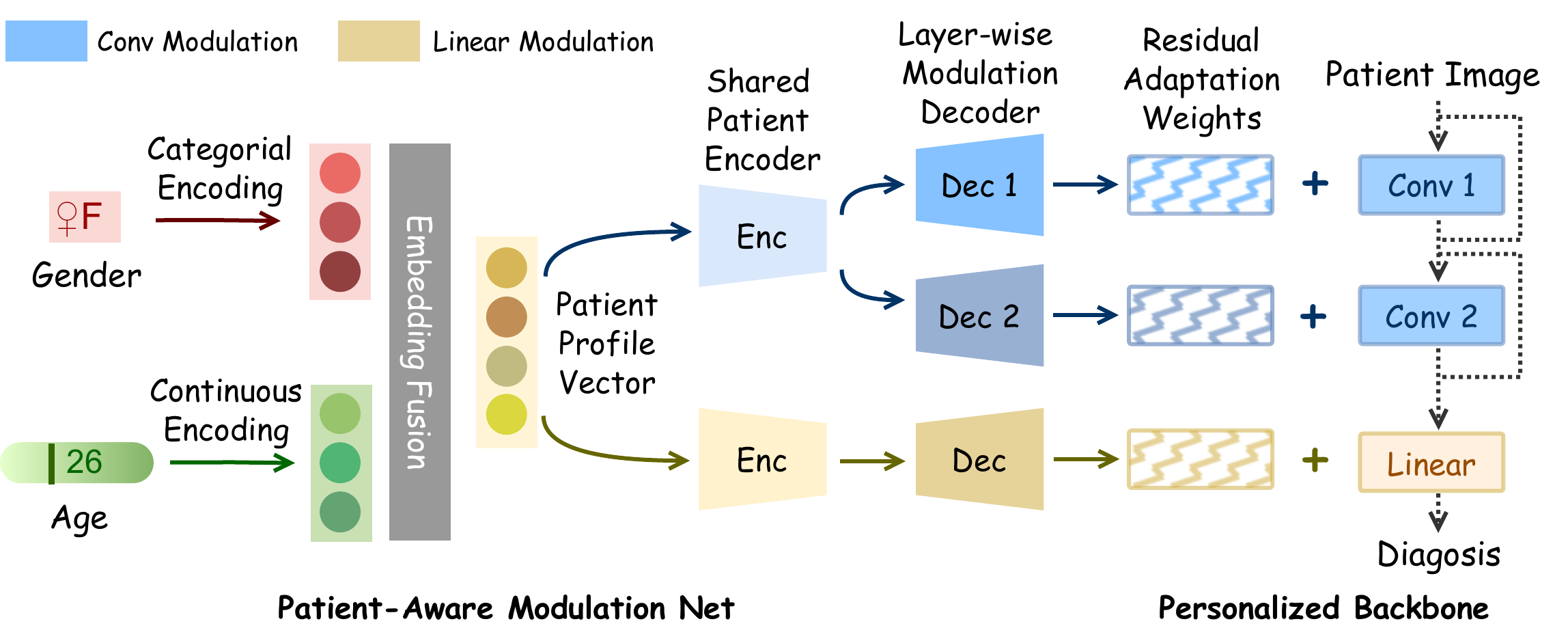}
  \caption{The HyperAdapt framework. Patient attributes are embedded and fused into a patient embedding, which is fed into hyper-adapter layers that generate subgroup-conditioned parameter offsets for selected convolutional and linear layers. These offsets modify the frozen base-model weights to produce adapted parameters used for inference. The adapted model then processes the input image to produce the diagnostic output.}
  \label{fig:framework}
\end{figure*}

\subsection{Problem Formulation}
We begin with a pretrained diagnostic model \(f: \mathcal{X} \rightarrow \mathcal{Y}\), parameterized by \(\theta\), that maps an input medical image \(x \in \mathcal{X}\) to a diagnostic output \(y \in \mathcal{Y}\). In practice, the accuracy of such models can vary substantially across patient subgroups, with performance typically higher for well-represented groups. This variability raises concerns regarding the reliability and equity of the model across diverse patient populations.
To address this limitation, we assume access to a vector of patient-specific attributes \(c_p\) for each patient \(p\). This vector includes demographic variables such as age or sex, as well as additional categorical clinical indicators such as summarized medical history or relevant genetic factors. Our goal is to adapt the general model \(f\) into an individualized model \(f_p\) whose behavior is conditioned on the attribute vector \(c_p\).

Formally, we aim to learn a hyper-adapter \(h(c_p; \phi)\), a neural network parameterized by \(\phi\), that generates patient-conditioned parameter offsets for the pretrained base model. Given the attribute vector \(c_p\), the hyper-adapter outputs a modification \(\Delta \theta_p\), which is applied to the base parameters to obtain adapted weights \(\theta_p = \theta + \Delta\theta_p\). The resulting patient-specific model \(f_p\) produces diagnostic outputs according to:
\[
    y_p = f(x;\, \theta + h(c_p;\phi)).
\]
This formulation enables parameter-level adaptation based on clinically relevant patient attributes, improving performance consistency across heterogeneous populations.

\subsection{Framework Overview}\label{sec:framework_overview}
Our framework performs conditional adaptation by generating patient-specific parameter modifications for a pretrained base model \(f(\cdot; \theta)\). For each patient \(p\), it receives an input medical image \(x\) together with an attribute vector \(c_p\) that encodes subgroup-relevant information.
As shown in Figure~\ref{fig:framework}, the adaptation procedure consists of three stages. First, the attribute vector \(c_p\) is processed by the adaptation module \(h(\cdot; \phi)\), which maps it into into a fixed-dimensional embedding representing the clinical characteristics captured by \(c_p\). Categorical and continuous attributes are embedded using separate transformations. Second, this embedding is passed through a set of parameter generators within \(h\) to produce the parameter offsets \(\Delta\theta_p\) for selected layers of the base model.
Third, the offsets are added to the frozen backbone parameters \(\theta\) to form the adapted weights $\theta_p = \theta + \Delta\theta_p$.
The resulting model \(f(\cdot; \theta_p)\) then processes the image \(x\) to produce the diagnostic prediction \(y_p\). Only the parameters \(\phi\) of the adaptation module are updated during training, and they are optimized end-to-end to minimize the diagnostic loss across the population.

The main technical challenge lies in designing the adaptation module \(h\). The module must map the low-dimensional attribute vector \(c_p\) to a high-dimensional offset vector \(\Delta\theta_p\), which can contain millions of parameters. This dimensionality gap creates a strong tendency toward overfitting, where the module may fit non-generalizable variations in the training subgroups instead of learning systematic patterns of patient-specific variability. In such cases, the generated offsets reflect memorized adjustments rather than stable, broadly applicable adaptations, which can degrade performance on unseen patients. The following sections present the architectural design choices and regularization mechanisms used to ensure that the generated offsets remain stable and effective.

\subsection{Embedding Subgroup-Relevant Attributes}

The first stage of the framework transforms the raw attribute vector \(c_p\) into a dense embedding suitable for subgroup-conditioned parameter adaptation. The vector may include both categorical and continuous variables, which differ in structure and therefore require distinct processing. Categorical variables represent discrete sets without inherent ordering, while continuous variables lie in an ordered numerical space that supports interpolation and generalization. Applying a single encoding mechanism to both can impose inappropriate geometric assumptions and obscure their statistical properties.

%such as continuous variables (e.g., age) and discrete variables (e.g., sex or coded clinical indicators).

To handle such differences, the embedding module \(h_{\text{embed}}\) processes the two attribute types through separate pathways: For categorical attributes, discrete variables are mapped using learnable embedding tables, assigning each category a distinct dense representation that reflects subgroup structure. For continuous attributes, numerical variables are processed by a small MLP. This pathway enables smooth interpolation between values not seen during training and avoids the abrupt transitions introduced by manual discretization. 
The outputs of these pathways are concatenated and passed through a fusion MLP to capture interactions between categorical and continuous indicators. This produces the unified embedding:
\[
    e_p = h_{\text{embed}}(c_p;\,\phi_{\text{embed}}),
\]
which serves as the input to the subsequent parameter-generation module.

Furthermore, as clinical data frequently contain missing entries, the framework incorporates mechanisms for handling incomplete attributes. For categorical variables, a dedicated learnable embedding vector is assigned to the missing category. For continuous variables, the model receives both a median-imputed value and a binary indicator that signals missingness. This design allows the model to remain robust to incomplete records while exploiting missingness patterns when they carry predictive value.

All parameters \(\phi_{\text{embed}}\) are optimized jointly with the rest of the system, ensuring that the learned embeddings are informative for generating subgroup-conditioned parameter offsets and improving performance across heterogeneous populations.

\subsection{Designing the Hyper-Adapter Layers}

After obtaining the patient embedding, the next step is to generate parameter offsets for the base model. Directly predicting a full weight modification for every layer in a modern network would make the adaptation module excessively large and prone to overfitting, given the low dimensionality of the attribute vector. Effective conditional adaptation therefore requires a compact and well-regularized mechanism for producing these offsets. In the following sections, we describe specialized low-rank adapters for linear and convolutional layers, which is the dominant layers in contemporary architectures. We then introduce a shared-parameter strategy that improves efficiency and generalization. %followed by a discussion on selecting which layers of the pre-trained model should receive adaptations. 

\subsubsection{Adapter for Linear Layers}
Consider a linear layer $\ell$ with the weight matrix $W_\ell \in \mathbb{R}^{d_{out} \times d_{in}}$. The goal is to generate a patient-specific modification without directly predicting the full weight tensor.  HyperAdapt instead produces two low-rank matrices, $A_{\ell,p} \in \mathbb{R}^{d_{out} \times k}$ and $B_{\ell,p} \in \mathbb{R}^{k \times d_{in}}$, where $k \ll d_{out}, d_{in}$. The resulting offset is reconstructed as $\Delta W_{\ell, p} = A_{\ell, p} B_{\ell, p}$.

This formulation reduces the number of generated parameters from $d_{out} \times d_{in}$ to $k \times (d_{out} + d_{in})$, yielding a more stable and data-efficient mapping from patient embeddings to layer-specific adaptations. The personalized weight is then given by $W_{\ell, p} = W_\ell + \Delta W_{\ell, p}$.

\subsubsection{Adapter for Convolutional Layers}

A convolutional layer $\ell$ has kernel weights $\Theta_\ell \in \mathbb{R}^{C_{out} \times C_{in} \times K_h \times K_w}$, where $C_{out}$ and $C_{in}$ denote the output and input channels, and $K_h, K_w$ are the spatial kernel dimensions. Directly adapting these high-dimensional tensors would again be prone to overfitting.

To preserve the spatial structure of convolutional kernels while enabling patient-conditioned adaptation, we adopt a channel-wise modulation strategy. The adapter generates two low-rank matrices, $A_{\ell,p} \in \mathbb{R}^{C_{out} \times k}$ and $B_{\ell,p} \in \mathbb{R}^{k \times C_{in}}$, which together define a modulation matrix $M_{\ell,p}^{i,j}$. The personalized kernel is obtained through multiplicative scaling:

\begin{equation*}
    \Theta_{\ell, p}^{i,j,:,:} = \Theta_\ell^{i,j,:,:} \cdot (1 + M_{\ell, p}^{i,j})
\end{equation*}

% We aim to learn a patient-specific modulation matrix $M_{\ell, p} \in \mathbb{R}^{C_{out} \times C_{in}}$, where each element $M_{\ell, p}^{i,j}$ is a scalar that re-weights the importance of the interaction between the $j$-th input channel and the $i$-th output channel.

% Therefore, the framework $h$ generates two low-rank matrices, $A_{\ell, p} \in \mathbb{R}^{C_{out} \times k}$ and $B_{\ell, p} \in \mathbb{R}^{k \times C_{in}}$. The full modulation matrix is then constructed as $M_{\ell, p} = A_{\ell, p} B_{\ell, p}$. The personalized kernel $\Theta_{\ell, p}$ is obtained by applying these learned channel-wise scalars multiplicatively:

This design efficiently adjusts inter-channel interactions while respecting the structured nature of convolutional filters and the coarse granularity of subgroup-level signals.

\subsubsection{Efficient Generation via Shared Generation Blocks}
Widely used architectures such as ResNet~\cite{he2016deep} contain many layers with identical input--output dimensionalities. 
Generating a separate pair $(A_{\ell,p}, B_{\ell,p})$ for every layer would introduce redundancy 
and substantially increase the size of adapters.

To exploit this repeated structure, we introduce \emph{shared-generation blocks}.  
For all linear layers with the same output dimension $d_{\text{out}}$, the adapter shares a single generator for the $A$-matrix
    $A_{\text{shared},p} \in \mathbb{R}^{d_{\text{out}} \times k}$.
Each layer retains its own $B_{\ell,p} \in \mathbb{R}^{k \times d_{\text{in},\ell}}$ to preserve 
layer-specific flexibility. The resulting offset is given by
\[
    \Delta W_{\ell,p} = A_{\text{shared},p} B_{\ell,p}, \qquad 
    \forall \ell \in \mathcal{L}_{d_{\text{out}}}.
\]

An analogous strategy is applied to convolutional adapters. All convolutional layers with the same output 
channel size $C_{\text{out}}$ share a single generator for the 
$A_{\ell,p} \in \mathbb{R}^{C_{\text{out}} \times k}$ component, while each layer maintains a 
layer-specific $B_{\ell,p}$.

This shared-generation design reduces the number of learnable parameters, improves computational efficiency, and provides implicit regularization. By encouraging layers with identical output dimensions to learn a common adaptation structure, the model stabilizes subgroup-conditioned parameter generation and improves generalization across the network.

%% file: tex/experiments.tex
\section{Experiments}
To evaluate the effectiveness of HyperAdapt, we conduct comprehensive experiments designed to answer the following research questions
(RQs):

\begin{itemize}
    \item \textbf{RQ1: Overall Diagnostic Performance.} How does the proposed framework compare with state-of-the-art baselines across different backbone architectures in terms of diagnostic accuracy?
    \item \textbf{RQ2: Group-wise Reliability.} Do the performance gains provided by the framework extend across patient demographic groups, and are the improvements particularly pronounced for underrepresented populations?
    \item \textbf{RQ3: Interpretability and Behavioral Consistency.} Does the framework learn a stable and clinically meaningful mapping from patient context to model behavior? 
    \item \textbf{RQ4: Ablation of Design Components.} What are the individual contributions of low-rank decomposition, channel-wise modulation, and shared-generation blocks to overall performance and parameter efficiency?
    \item \textbf{RQ5: Resource-performance Trade-off.} How does diagnostic performance vary with the capacity of the adaptive module, such as the depth of the generator or the rank $k$ used in the low-rank adapters?
\end{itemize}

\input{table/resnet50}
\input{table/swin-t}
\subsubsection{Datasets}
To evaluate the proposed method under diverse subgroup conditions, experiments are conducted on three publicly available medical imaging datasets. The Fitzpatrick-17k dataset~\cite{groh2021evaluating} contains 16,577 skin lesion images across 114 dermatological diagnoses. Each image is labeled with a Fitzpatrick skin type from one to six, which serves as the subgroup attribute. The ODIR-5k dataset~\cite{ODIR5Kdataset} provides color fundus photographs from 5,000 patients and includes annotations for eight diagnostic categories. Each image is accompanied by age and gender information, enabling subgroup analysis along multiple demographic dimensions. The PAD-UFES-20 dataset~\cite{pacheco2020pad} consists of 2,298 skin lesion images associated with six primary lesion categories and includes extensive clinical metadata. Up to 26 attributes are available for each case, covering demographic information, lesion characteristics, and lesion diameter.
All images are resized to 128×128 pixels and augmented with random horizontal and vertical flips, rotation, scaling, and AutoAugment~\cite{cubuk2018autoaugment}. Each dataset is partitioned into training, validation, and test sets using a ratio of 6:2:2. 

\subsubsection{Implementation Details}
Unless otherwise specified, ResNet-50~\cite{he2016deep} is adopted as the backbone architecture. We integrate the hyper-adapter modules into all convolutional stages of the network except the initial stem block, which consists of the first convolution and max-pooling layers. In ResNet-50, the stem is followed by four sequential residual stages (conv2\_x through conv5\_x); each of these stages receives a hyper-adapter extension that generates structured offset weights for the corresponding convolutional layers. In addition to the convolutional hierarchy, the final fully connected classification layer also receives a hyper-adapter extension, allowing instance-conditioned adjustments at the output level.
All models are trained for 200 epochs using the Adam optimizer, with a fixed batch size of 128 to ensure consistency across methods. For the learning rate, we performed a small validation-based search over $\{10^{-2}, 10^{-3}, 10^{-4}\}$ and selected $10^{-3}$, which yielded the best performance. The initial learning rate decays by a factor of 10 every 100 epochs. These training settings are applied uniformly to both our proposed methods and all baseline models. To account for variability, each experiment is repeated three times, and the average performance is reported.

% \input{table/dataset}
% A summary of the dataset characteristics is provided in Table~\ref{tab:dataset}. All datasets are preprocessed using standard procedures, including resizing and data augmentation [1]. Each dataset is divided into training, validation, and test sets with a split ratio of 6:2:2.

\subsubsection{Baselines}
We compare HyperAdapt with three categories of baselines that reflect the major strategies for improving subgroup performance. Because subgroup reliability intersects with fairness optimization, group-specific modeling, and metadata integration, the evaluation includes representative methods from each direction. Fairness-oriented baselines FairAdaBN~\cite{xu2023fairadabn} and FairQuantize~\cite{guo2024fairquantize} reduce disparities across demographic groups through adaptive normalization or quantization. Group-specific methods  GroupModel~\cite{puyol2021fairness} and GroupAdapt~\cite{guo2021parameter} allocate dedicated model capacity to each subgroup via separate models or group-conditioned parameters. Metadata-fusion approaches, DAFT~\cite{wolf2022daft} and HEAL~\cite{hemker2024healnet}, incorporate demographic or clinical attributes directly into the feature extraction process. Together, these baselines cover the primary methodological directions in subgroup-aware modeling, providing a comprehensive context for evaluating the subgroup robustness achieved by HyperAdapt.

\begin{figure*}[t]
  \centering
  \includegraphics[width=\textwidth]{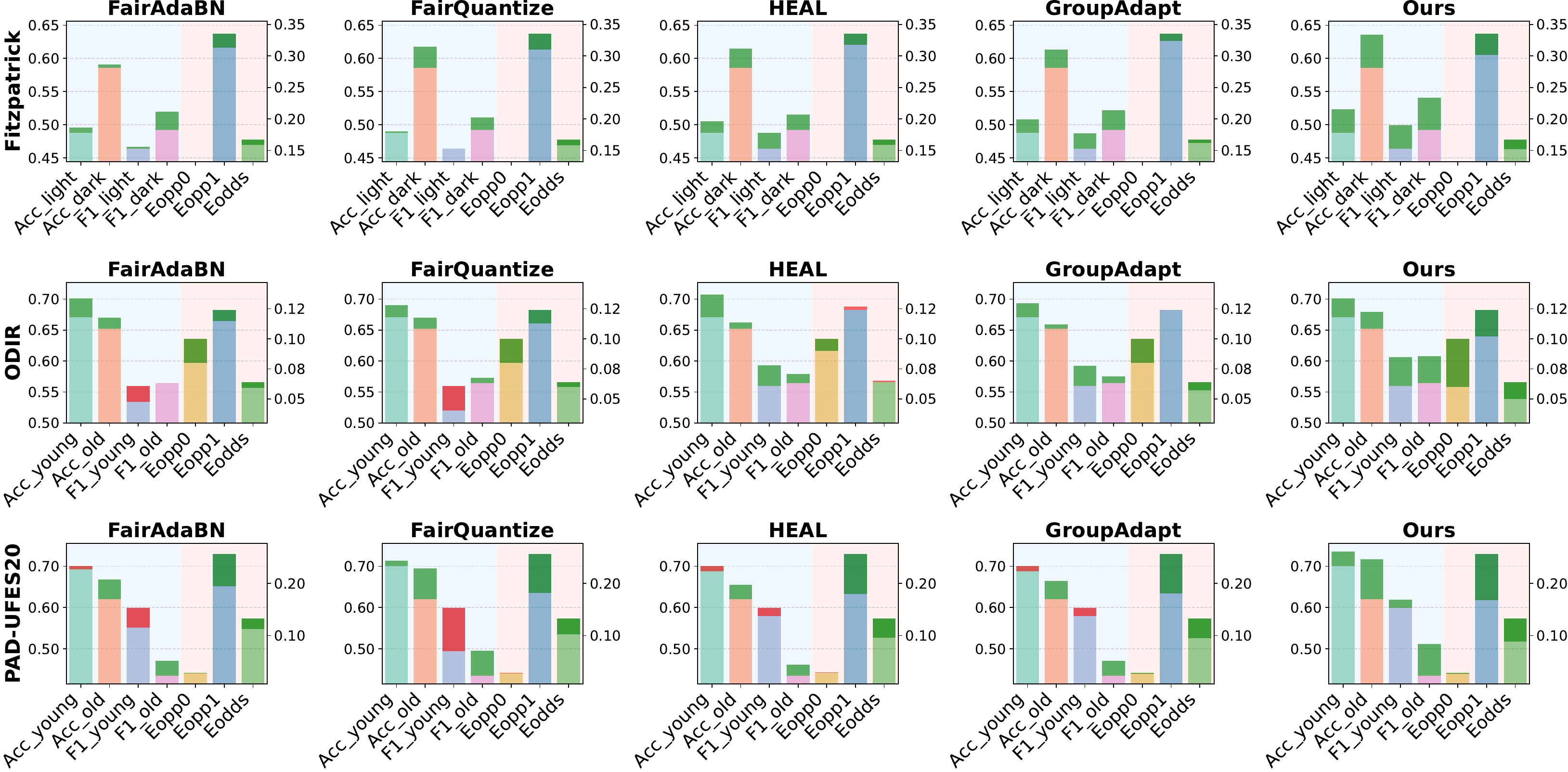}
  \caption{Group-wise performance comparison across Fitzpatrick-17k, ODIR-5k, and PAD-UFES-20. Each panel shows the change in accuracy, F1, and fairness metrics (Eopp0, Eopp1, Eodds) for each method relative to the vanilla baseline. Green bars denote performance improvements, and red bars denote declines.}
  \label{fig:group_diff}
\end{figure*}

\subsection{Experiment Result}
\subsubsection{RQ1: Overall Diagnostic Performance}

\textbf{Result on ResNet50 backbone}
Across the three datasets evaluated with ResNet-50 as the backbone, HyperAdapt consistently achieves the best overall performance. On Fitzpatrick-17k, the proposed method reaches an accuracy of 0.563 and an F1 score of 0.538, improving upon the strongest baseline, GroupAdapt, by 1.7\% and 1.6\% in both metrics. These gains indicate more reliable handling of skin-type-related distribution shifts.
On ODIR-5k, performance differences among methods are smaller, yet HyperAdapt still attains the highest F1 score at 0.613, exceeding HEAL by 1.2\% and GroupAdapt by 1.8\%. In addition, it achieves the best recall at 0.598, which is especially relevant for medical screening tasks that prioritize sensitivity.
The largest improvements are observed on PAD-UFES-20, where HyperAdapt attains an accuracy of 0.726 and an F1 score of 0.644. This represents an absolute improvement of over 4\% over GroupAdapt and HEAL across both metrics. The pronounced gains on this dataset suggest that conditional adaptation becomes increasingly beneficial when richer clinical metadata are available.
Overall, these results demonstrate that HyperAdapt delivers consistent and meaningful improvements across diverse medical imaging benchmarks, with particularly strong benefits on smaller and noisier datasets where robust subgroup-aware modeling is critical.

%To examine the generality of our approach across backbone architectures, we further evaluate HyperAdapt using Swin-T, a transformer-based backbone widely adopted in medical imaging. All adaptive modules are applied to the linear layers of the transformer. Under this setting, HyperAdapt again yields consistent improvements over all baselines. On Fitzpatrick-17k, HyperAdapt achieves an accuracy of 0.635 and an F1 score of 0.621, improving upon the strongest baseline, GroupAdapt, by 0.9\% in accuracy and 1.3\% in F1. These gains indicate that the proposed adaptation mechanism remains effective beyond convolutional architectures. On ODIR-5k, HyperAdapt further advances performance, reaching an F1 score of 0.656, which exceeds HEAL by 2.4\% and GroupAdapt by 2.6\%. The recall improves to 0.641, representing a clear margin over all competing methods and underscoring its relevance for ophthalmic screening scenarios where sensitivity is critical. The largest improvements are observed on PAD-UFES-20. HyperAdapt attains an accuracy of 0.763 and an F1 score of 0.727, surpassing GroupAdapt by 1.7\% in accuracy and 3.3\% in F1, and exceeding HEAL by 2.8\% in accuracy and 2.1\% in F1. These results demonstrate the robustness of HyperAdapt when modeling high-dimensional clinical attributes. Overall, the results confirm that HyperAdapt generalizes effectively to transformer-based backbones and consistently delivers improved subgroup-aware performance across diverse datasets.

\textbf{Result on Different backbone }To assess generalization beyond convolutional architectures, we evaluate HyperAdapt using Swin-T~\cite{liu2021swin}, with adaptive modules applied to the linear layers. Under this setting, HyperAdapt consistently outperforms all baselines across datasets.
On Fitzpatrick-17k, HyperAdapt achieves an accuracy of 0.635 and an F1 score of 0.621, improving over GroupAdapt by 0.9\% in accuracy and 1.3\% in F1. On ODIR-5k, it attains the highest F1 score at 0.656, exceeding HEAL and GroupAdapt by approximately 2.5\%, and achieves the best recall at 0.641, which is important for ophthalmic screening scenarios.
The largest gains are observed on PAD-UFES-20, where HyperAdapt reaches an accuracy of 0.763 and an F1 score of 0.727. This corresponds to improvements of 2.1\% in F1 and 1.7\% in accuracy compared to the strongest baselines.
Overall, these results show that HyperAdapt generalizes well to transformer-based backbones and provides consistent subgroup-aware performance gains across datasets, with particularly strong benefits when modeling high-dimensional clinical attributes.

\subsubsection{RQ2: Group-wise Reliability}
Following the overall performance results, we analyze whether the observed gains are consistent across different subgroups. Figure~\ref{fig:group_diff} reports group-wise changes on Fitzpatrick-17k, ODIR-5k, and PAD-UFES-20 relative to the vanilla baseline, covering both performance metrics (Accuracy and F1) and fairness metrics (Eopp0, Eopp1, and Eodds, with lower values indicating smaller disparities~\cite{hardt2016equality, xu2025incorporating}). Performance gains and declines are visualized as positive (\textcolor{green!60!black}{green}) and negative (\textcolor{red!70!black}{red}) bars, respectively.
Across all datasets, HyperAdapt yields more uniform improvements across demographic groups than competing methods, with larger gains concentrated on historically underperforming groups. On Fitzpatrick-17k, HyperAdapt increases F1 for light-skin groups by 3.5\% and for dark-skin groups by 4.9\%, indicating a stronger benefit for disadvantaged populations. Similar patterns are observed on ODIR-5k and PAD-UFES-20, where improvements are both larger in magnitude and more evenly distributed across groups.

HyperAdapt also consistently reduces group disparities, as indicated by the fairness metrics. On Fitzpatrick-17k, Eopp1 decreases by 33.3\% and Eodds by 9.0\% relative to the vanilla baseline. On ODIR-5k, Eopp1 is reduced by 17.7\% and Eodds by 21.9\%, while on PAD-UFES-20 the reductions reach 34.6\% for Eopp1 and 33.4\% for Eodds. These results show that the proposed method improves subgroup-level performance while reducing group disparities, even without introducing explicit fairness constraints. This may because HyperAdapt enables group-conditioned parameter adaptation, allowing the model to better accommodate group-specific patterns rather than over-optimizing for any single group.
%Overall, HyperAdapt achieves consistent gains across groups and datasets, demonstrating improved alignment between predictive performance and subgroup reliability.

\begin{figure*}[t]
  \centering
  \includegraphics[width=0.9\textwidth]{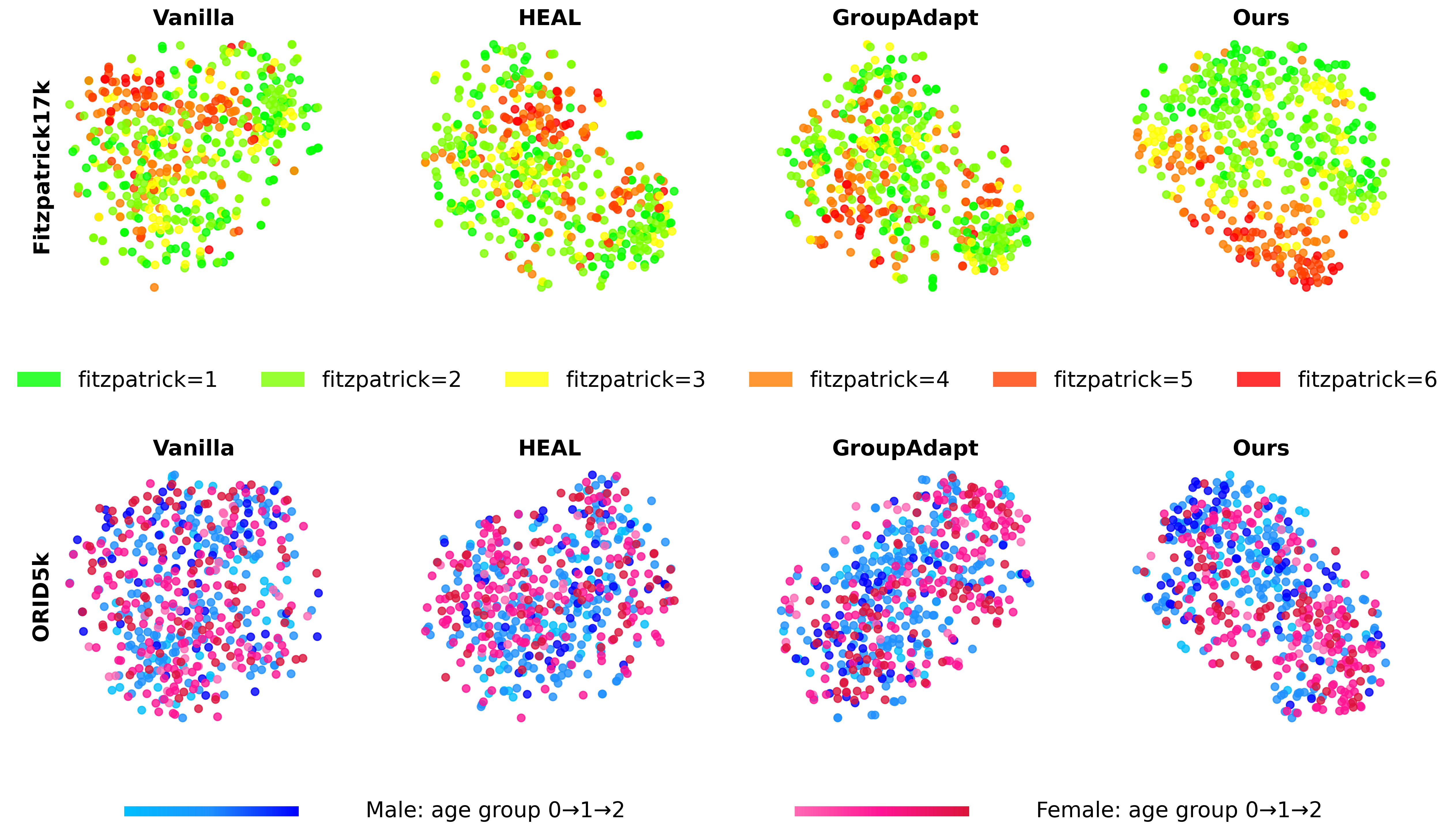}
  \caption{t-SNE Visualization of Contextual Feature Space Organization. The plots compare the feature space (from a single disease class) learned by different methods. (Top Row) On Fitzpatrick17k, features are colored by the 6 ordinal skin types. (Bottom Row) On ORID5k, features are colored by composite gender and age groups.}
  \vspace{-8pt}
  \label{fig:tsne}
\end{figure*}

\begin{figure*}[t]
  \centering
  \includegraphics[width=\textwidth]{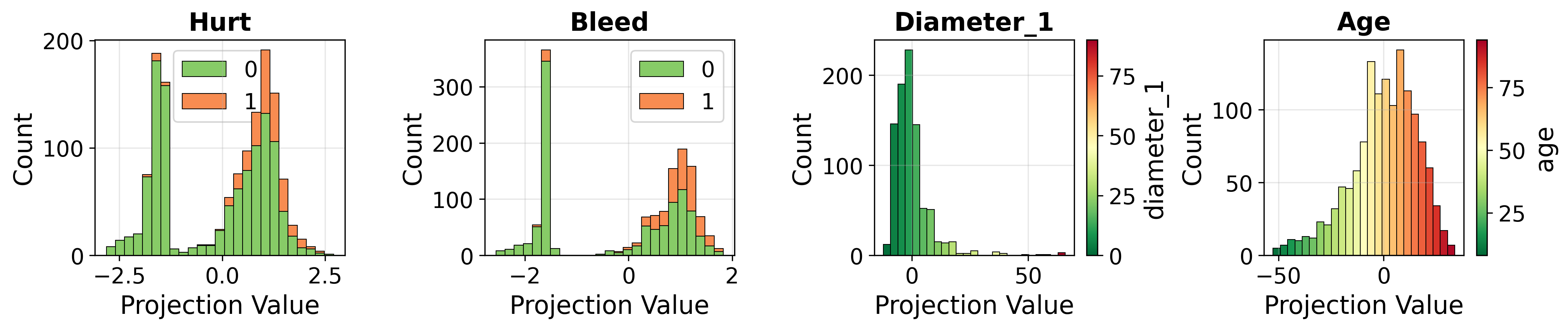}
  \caption{Linear probing visualizations of selected metadata attributes on PAD-UFES-20. These visualizations illustrate how clinically relevant metadata manifests structured alignment within the learned image embedding space.}
  \label{fig:pad}
  \vspace{-8pt}
\end{figure*}

\subsubsection{RQ3: Interpretability and Behavioral Consistency.}

To investigate the interpretability and consistency of our framework, we analyze the learned feature space using t-SNE to examine whether the model captures meaningful patient context and its functional role in representation learning. To isolate the effect of patient context and avoid confounding from the primary diagnostic label, this analysis is conducted on samples drawn from the class with the largest number of instances. We report results on Fitzpatrick-17k, which involves a one-dimensional ordinal context, and ORID-5k, which involves a two-dimensional composite context. We exclude PAD-UFES-20 from this visualization, as its patient context is derived from over 20 dimensions, making reliable 2D projection and structural interpretation infeasible.

As shown in Figure~\ref{fig:tsne}, the features from different groups in Vanilla and HEAL are largely intermixed. GroupAdapt demonstrates partial organization: on ORID-5k, it separates samples by gender but fails to capture age within each gender cluster.
In contrast, our framework learns a more coherent and structured feature representation. On Fitzpatrick-17k, it recovers the continuous ordinal progression of skin types, forming a smooth manifold that transitions from Fitzpatrick-1 to Fitzpatrick-6. This indicates that the model captures relational structure among groups rather than treating them as independent categories. More importantly, on the multi-dimensional ORID-5k context, our method achieves a hierarchical organization: it first separates samples into clear gender-specific clusters and then preserves a smooth age-related gradient within each cluster. This two-level structure demonstrates our framework’s ability to map complex, composite patient profiles into predictable and well-organized feature adaptations, confirming a meaningful and consistent correspondence between patient context and model behavior.

PAD-UFES-20 provides a more heterogeneous set of metadata than Fitzpatrick-17k and ODIR-5K, including continuous clinical measurements (e.g., lesion diameter and patient age), discrete symptom indicators (e.g., pain and bleeding), and higher-cardinality background attributes. We perform a linear probing analysis on this dataset to assess whether individual metadata attributes are encoded in an interpretable manner within the learned image embeddings. For each selected representative attribute, we train a linear probe to identify a projection direction in the embedding space and project all samples onto the resulting one-dimensional axis. The projected distributions are visualized using class-conditional histograms for discrete attributes and value-colored histograms for continuous attributes. Figure~\ref{fig:pad} shows representative results for two symptom-related attributes (hurt and bleed) and two continuous attributes (diameter and age). Symptom-related attributes exhibit clear separation trends, with positive cases concentrating in higher projection regions despite class imbalance. Continuous attributes display smooth, approximately monotonic variation along the projection axis. These results indicate that the embedding encodes clinically relevant metadata in a structured and interpretable manner, even under high-dimensional attribute settings.

\input{table/ablation_study}
\subsubsection{RQ4: Ablation of Design Components.}
To assess the contribution of each design component, we conduct an ablation study summarized in Table~\ref{tab:ablation}, isolating the effects of channel-wise modulation, low-rank decomposition, and shared generation blocks on both diagnostic performance and parameter efficiency.
We begin with a baseline that directly generates dense parameters. This configuration exhibits degraded accuracy and F1 score while incurring a substantial parameter cost of 184M, indicating that unconstrained dense parameter generation is both inefficient and ineffective. Introducing channel-wise modulation leads to a marked improvement in both performance and efficiency: the parameter count is reduced to approximately 4M, while accuracy increases to 0.546. This result highlights channel-level adaptation as a strong and parameter-efficient form of conditional modeling.
Incorporating low-rank decomposition further improves performance. With both activation modulation and low-rank weight updates, accuracy increases to 0.557 and F1 to 0.532, while the parameter count decreases to 3.1M. These gains suggest complementary benefits from jointly adapting activations and weights. Finally, introducing shared generation blocks yields the most compact and best-performing configuration. The resulting model achieves an accuracy of 0.563 and an F1 score of 0.538 with only 2.5M parameters, representing the highest performance under the lowest parameter budget. Overall, these results indicate that sharing generator components serves as an effective regularization mechanism, improving both predictive performance and model compactness.

% \begin{figure}[htbp]
%   \centering
%   \includegraphics[width=0.2\textwidth]{fig/figure_lora.pdf}
%   \caption{tttt}
%   \label{fig:lora}
% \end{figure}

\begin{figure}[htbp]
    \centering
    \begin{subfigure}{0.492\columnwidth}
        \centering
        \includegraphics[width=\linewidth]{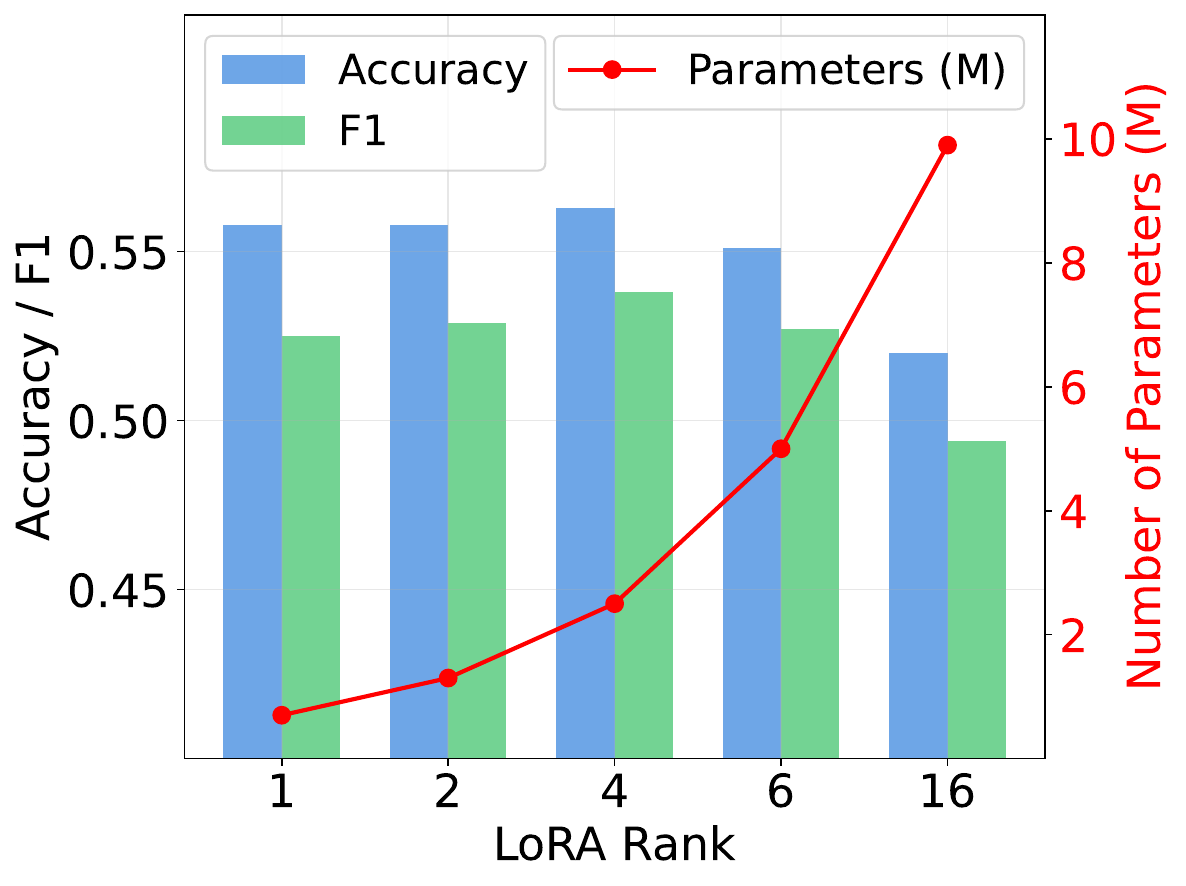}
        \caption{LoRA Illustration}
        \label{fig:lora}
    \end{subfigure}
    %\hfill
    \begin{subfigure}{0.492\columnwidth}
        \centering
        \includegraphics[width=\linewidth]{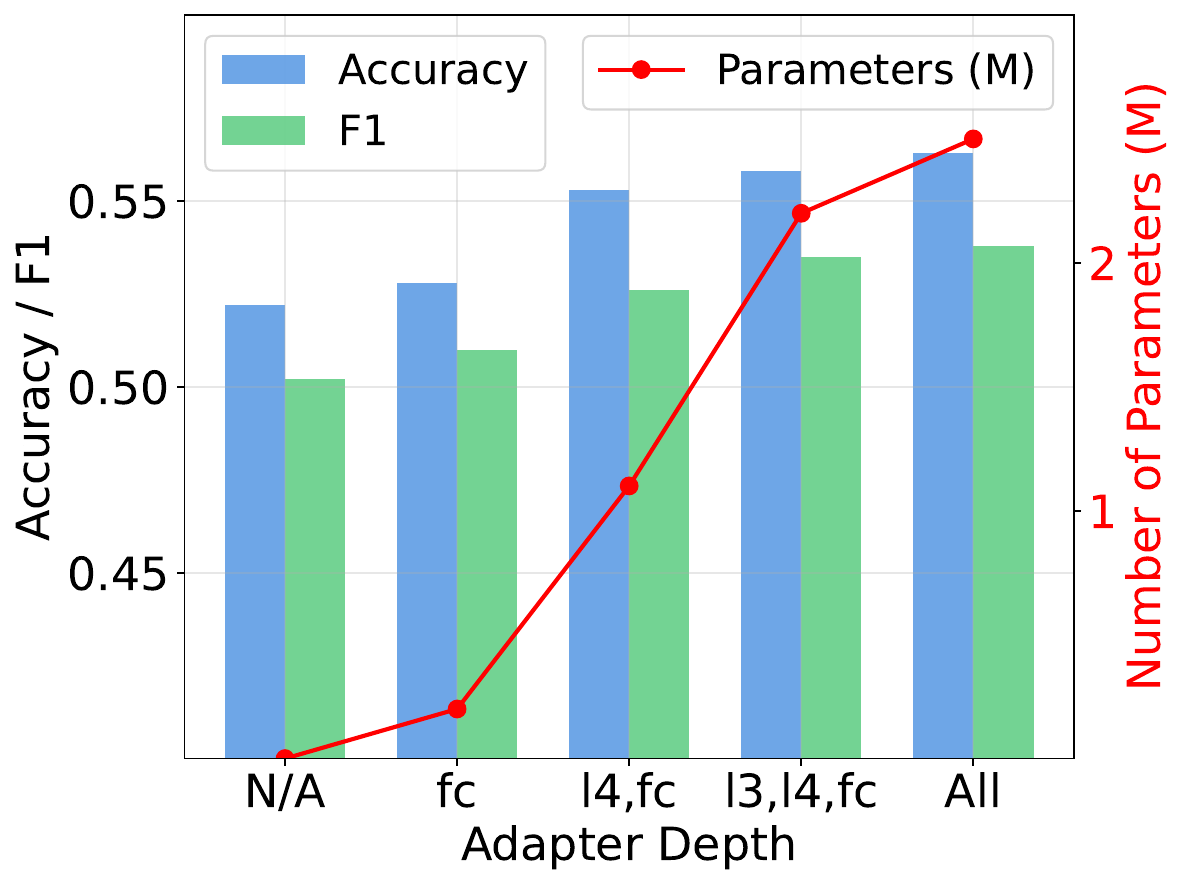}
        \caption{Depth Illustration}
        \label{fig:depth}
    \end{subfigure}
    \caption{Comparison of LoRA and Depth modules.}
    \label{fig:lora_depth}
\end{figure}

\subsubsection{RQ5: Resource-performance Trade-off}
In this section, we analyze how additional model capacity influences diagnostic performance by examining the trade-off between accuracy and F1 score relative to the number of parameters introduced by adaptive modules, as summarized in Figure~\ref{fig:lora_depth}. The analysis focuses on the LoRA rank and the depth at which adapters are integrated into the network.

\textbf{LoRA Rank.}
Increasing the LoRA rank expands modulation capacity at the cost of additional parameters. Performance improves from rank~1 to rank~4, where accuracy reaches 0.563 and F1 reaches 0.538 with 2.5M parameters. Further increases provide limited benefit. Rank~6 adds roughly 5M parameters without improving performance, while rank~16 enlarges the model to about 9.9M parameters and reduces accuracy to 0.520 and F1 to 0.494. These results indicate that gains from low-rank modulation saturate quickly and that overly large ranks can be counterproductive.

\textbf{Adapter Depth.}
We further examine the effect of distributing adaptive modules across network depth. Adapting only the final classification layer yields modest improvements. Extending adaptation to include the fourth convolutional stage increases accuracy from 0.553 to 0.558 with an additional 1.1M parameters. The best performance is achieved when all stages, from the deepest convolutional layers to the final classifier, are adapted, reaching accuracy 0.563 and F1 0.538 with 2.5M parameters. This configuration outperforms shallower designs while remaining substantially more parameter-efficient than high-rank alternatives.

%Overall, the results identify a clear resource–performance frontier. Moderate LoRA ranks combined with deeper adapter integration provide the most efficient performance gains, whereas excessively large ranks or shallow integration lead to diminishing returns.

%% file: table/resnet50.tex
\begin{table*}[ht]
\centering
\scriptsize
\caption{Performance comparison between our method and baseline methods in terms of medical image classification metrics using ResNet50 as backbone}
\begin{tabular}{l|cccc|cccc|cccc}
\toprule
\multirow{2}{*}{Method} 
& \multicolumn{4}{c|}{Fitzpatrick-17k} 
& \multicolumn{4}{c|}{ODIR-5k}       
& \multicolumn{4}{c}{PAD-UFES-20}   \\
& ACC & PR & RC & F1
& ACC & PR & RC & F1
& ACC & PR & RC & F1 \\ \midrule

Vanilla
& 0.522 & 0.531 & 0.508 & 0.502
& 0.650 & 0.642 & 0.556 & 0.576
& 0.661 & 0.641 & 0.574& 0.596\\

FairAdaBN
& 0.528 & 0.512 & 0.518 & 0.499
& \underline{0.669} & 0.646 & 0.567 & 0.579
& 0.680 & 0.627 & 0.568 & 0.588 \\

FairQuantize
& 0.534 & 0.546 & 0.505 & 0.508
& 0.662 & \textbf{0.665} & 0.562 & 0.580
& \underline{0.704} & \underline{0.668} & 0.571 & 0.597 \\

GroupModel
& 0.536 & 0.547 & 0.506 & 0.507
& \textbf{0.672} & \underline{0.662} & 0.573 & 0.586
& 0.687 & 0.649 & 0.581 & 0.595 \\

DAFT
& 0.537 & 0.542 & 0.512 & 0.511
& 0.663 & 0.648 & \underline{0.585} & 0.598
& 0.654 & 0.645 & \underline{0.584} & 0.598 \\

HEAL
& 0.542 & 0.558 & 0.516 & 0.521
& 0.662 & 0.656 & 0.584 & \underline{0.601}
& 0.672 & 0.615 & 0.580 & 0.594 \\

GroupAdapt
& \underline{0.546} & \underline{0.562} & \underline{0.520} & \underline{0.522}
& 0.667 & 0.656 & 0.579 & 0.595
& 0.676 & 0.628 & 0.583 & \underline{0.600} \\

Ours
& \textbf{0.563} & \textbf{0.576} & \textbf{0.536} & \textbf{0.538}
& \underline{0.669} & 0.661 & \textbf{0.598} & \textbf{0.613}
& \textbf{0.726} & \textbf{0.681} & \textbf{0.625} & \textbf{0.644} \\

\bottomrule
\end{tabular}
\end{table*}

%% file: table/swin-t.tex
\begin{table*}[htbp]
\centering
\scriptsize
\caption{Performance comparison between our method and baseline methods in terms of medical image classification metrics using Swin-T as backbone}
\begin{tabular}{l|cccc|cccc|cccc}
\toprule
\multirow{2}{*}{Method} 
& \multicolumn{4}{c|}{Fitzpatrick-17k} 
& \multicolumn{4}{c|}{ODIR-5k}       
& \multicolumn{4}{c}{PAD-UFES-20}   \\
& ACC & PR & RC & F1 
& ACC & PR & RC & F1 
& ACC & PR & RC & F1 \\ \midrule

Vanilla      
& 0.607 & 0.625 & 0.587 & 0.587  
& 0.676 & 0.659 & 0.619 & 0.614  
& 0.754 & 0.750 & 0.662 & 0.692  \\

FairAdaBN         
& \underline{0.633} & 0.636 & 0.582 & 0.587  
& 0.655 & 0.641 & 0.618 & 0.616  
& 0.748 & 0.723 & 0.679 & 0.685  \\ 

FairQuantize         
& 0.632 & 0.619 & 0.591 & 0.588  
& \textbf{0.696} & \textbf{0.680} & 0.576 & 0.598  
& \underline{0.759} & 0.746 & 0.678 & 0.697  \\

GroupModel       
& 0.608 & 0.630 & 0.590 & 0.593  
& 0.686 & 0.668 & 0.619 & 0.599  
& 0.754 & 0.729 & 0.673 & 0.694  \\

DAFT  
& 0.621 & 0.639 & 0.604 & 0.603  
& 0.685 & 0.676 & 0.610 & 0.628  
& 0.733 & 0.721 & 0.656 & 0.681  \\

HEAL         
& 0.622 & 0.650 & \underline{0.610} & \underline{0.609}  
& 0.680 & 0.670 & \underline{0.626} & \underline{0.632}  
& 0.735 & 0.751 & \underline{0.684} & \underline{0.706}  \\

GroupAdapt   
& 0.626 & \underline{0.654} & 0.603 & 0.608  
& 0.687 & 0.677 & 0.618 & 0.630  
& 0.746 & \underline{0.753} & 0.659 & 0.694  \\

Ours         
& \textbf{0.635} & \textbf{0.666} & \textbf{0.621} & \textbf{0.621}  
& \underline{0.690} & \underline{0.679} & \textbf{0.641} & \textbf{0.656}  
& \textbf{0.763} & \textbf{0.788} & \textbf{0.691} & \textbf{0.727}  \\

\bottomrule
\end{tabular}
\end{table*}

%% file: table/ablation_study.tex
\begin{table}[htbp]
\caption{Ablation Study.}
\label{tab:ablation}
\begin{tabular}{@{}ccc|ccc@{}}
\toprule
Channel-wise & LoRA & Sharing & Accuracy & F1    & \# Parameters \\ \midrule
      &      &         & 0.518    & 0.492 & 184M      \\
\checkmark     &      &         & 0.546    & 0.515 & 4M       \\
\checkmark     & \checkmark    &         & 0.557    & 0.532 & 3.1M       \\
\checkmark     & \checkmark    & \checkmark       & 0.563    & 0.538 & 2.5M       \\ \bottomrule
\end{tabular}
\end{table}

%% file: tex/conclusion.tex
\section{Conclusion}
This work focuses on achieving reliable diagnostic performance across diverse patient populations while retaining the clinically meaningful information contained in patient attributes. We present a patient-conditioned adaptation framework that allows pretrained backbones to adjust their computations based on patient attributes through lightweight residual parameter updates. The framework incorporates low-rank and shared-parameter constraints to ensure efficiency and maintain the advantages of pretrained initialization. The empirical results further demonstrate that modeling patient-dependent variability can improve subgroup-level robustness while preserving overall accuracy, pointing toward a promising direction for building medical-AI systems that are both effective and equitable in real-world deployment.

%% file: ref.bib
@article{kline2022multimodal,
  title={Multimodal machine learning in precision health: A scoping review},
  author={Kline, Adrienne and Wang, Hanyin and Li, Yikuan and Dennis, Saya and Hutch, Meghan and Xu, Zhenxing and Wang, Fei and Cheng, Feixiong and Luo, Yuan},
  journal={NPJ digital medicine},
  volume={5},
  number={1},
  pages={171},
  year={2022},
  publisher={Nature Publishing Group UK London}
}

@article{baltruvsaitis2018multimodal,
  title={Multimodal machine learning: A survey and taxonomy},
  author={Baltru{\v{s}}aitis, Tadas and Ahuja, Chaitanya and Morency, Louis-Philippe},
  journal={IEEE transactions on pattern analysis and machine intelligence},
  volume={41},
  number={2},
  pages={423--443},
  year={2018},
  publisher={IEEE}
}

@inproceedings{perez2018film,
  title={Film: Visual reasoning with a general conditioning layer},
  author={Perez, Ethan and Strub, Florian and De Vries, Harm and Dumoulin, Vincent and Courville, Aaron},
  booktitle={Proceedings of the AAAI conference on artificial intelligence},
  volume={32},
  number={1},
  year={2018}
}

@article{wolf2022daft,
  title={DAFT: A universal module to interweave tabular data and 3D images in CNNs},
  author={Wolf, Tom Nuno and P{\"o}lsterl, Sebastian and Wachinger, Christian and Alzheimer’s Disease Neuroimaging Initiative and others},
  journal={NeuroImage},
  volume={260},
  pages={119505},
  year={2022},
  publisher={Elsevier}
}

@article{ha2016hypernetworks,
  title={Hypernetworks},
  author={Ha, David and Dai, Andrew and Le, Quoc V},
  journal={arXiv preprint arXiv:1609.09106},
  year={2016}
}

@article{jia2016dynamic,
  title={Dynamic filter networks},
  author={Jia, Xu and De Brabandere, Bert and Tuytelaars, Tinne and Gool, Luc V},
  journal={Advances in neural information processing systems},
  volume={29},
  year={2016}
}

@inproceedings{aharon2023hypernetwork,
  title={Hypernetwork-based adaptive image restoration},
  author={Aharon, Shai and Ben-Artzi, Gil},
  booktitle={ICASSP 2023-2023 IEEE International Conference on Acoustics, Speech and Signal Processing (ICASSP)},
  pages={1--5},
  year={2023},
  organization={IEEE}
}

@inproceedings{littwin2019deep,
  title={Deep meta functionals for shape representation},
  author={Littwin, Gidi and Wolf, Lior},
  booktitle={Proceedings of the IEEE/CVF international conference on computer vision},
  pages={1824--1833},
  year={2019}
}

@article{von2019continual,
  title={Continual learning with hypernetworks},
  author={Von Oswald, Johannes and Henning, Christian and Grewe, Benjamin F and Sacramento, Jo{\~a}o},
  journal={arXiv preprint arXiv:1906.00695},
  year={2019}
}

@inproceedings{shamsian2021personalized,
  title={Personalized federated learning using hypernetworks},
  author={Shamsian, Aviv and Navon, Aviv and Fetaya, Ethan and Chechik, Gal},
  booktitle={International conference on machine learning},
  pages={9489--9502},
  year={2021},
  organization={PMLR}
}

@inproceedings{wydmanski2023hypertab,
  title={Hypertab: Hypernetwork approach for deep learning on small tabular datasets},
  author={Wydma{\'n}ski, Witold and Bulenok, Oleksii and {\'S}mieja, Marek},
  booktitle={2023 IEEE 10th International Conference on Data Science and Advanced Analytics (DSAA)},
  pages={1--9},
  year={2023},
  organization={IEEE}
}

@inproceedings{zhang2018mitigating,
  title={Mitigating unwanted biases with adversarial learning},
  author={Zhang, Brian Hu and Lemoine, Blake and Mitchell, Margaret},
  booktitle={Proceedings of the 2018 AAAI/ACM Conference on AI, Ethics, and Society},
  pages={335--340},
  year={2018}
}

@inproceedings{wu2022fairprune,
  title={Fairprune: Achieving fairness through pruning for dermatological disease diagnosis},
  author={Wu, Yawen and Zeng, Dewen and Xu, Xiaowei and Shi, Yiyu and Hu, Jingtong},
  booktitle={International Conference on Medical Image Computing and Computer-Assisted Intervention},
  pages={743--753},
  year={2022},
  organization={Springer}
}

@misc{sabuncu2025ethical,
  title={Ethical use of artificial intelligence in medical diagnostics demands a focus on accuracy, not fairness},
  author={Sabuncu, Mert R and Wang, Alan Q and Nguyen, Minh},
  journal={NEJM AI},
  volume={2},
  number={1},
  pages={AIp2400672},
  year={2025},
  publisher={Massachusetts Medical Society}
}

@inproceedings{xu2025incorporating,
  title={Incorporating Rather Than Eliminating: Achieving Fairness for Skin Disease Diagnosis Through Group-Specific Experts},
  author={Xu, Gelei and Duan, Yuying and Liu, Zheyuan and Li, Xueyang and Jiang, Meng and Lemmon, Michael and Jin, Wei and Shi, Yiyu},
  booktitle={International Conference on Medical Image Computing and Computer-Assisted Intervention},
  pages={284--294},
  year={2025},
  organization={Springer}
}

@article{caini2009meta,
  title={Meta-analysis of risk factors for cutaneous melanoma according to anatomical site and clinico-pathological variant},
  author={Caini, Saverio and Gandini, Sara and Sera, Francesco and Raimondi, Sara and Fargnoli, Maria Concetta and Boniol, Mathieu and Armstrong, Bruce K},
  journal={European journal of cancer},
  volume={45},
  number={17},
  pages={3054--3063},
  year={2009},
  publisher={Elsevier}
}

@article{narayanan2010ultraviolet,
  title={Ultraviolet radiation and skin cancer},
  author={Narayanan, Deevya L and Saladi, Rao N and Fox, Joshua L},
  journal={International journal of dermatology},
  volume={49},
  number={9},
  pages={978--986},
  year={2010},
  publisher={Wiley Online Library}
}

@inproceedings{gordon2013skin,
  title={Skin cancer: an overview of epidemiology and risk factors},
  author={Gordon, Randy},
  booktitle={Seminars in oncology nursing},
  volume={29},
  number={3},
  pages={160--169},
  year={2013},
  organization={Elsevier}
}

@inproceedings{guo2024fairquantize,
  title={FairQuantize: Achieving Fairness Through Weight Quantization for Dermatological Disease Diagnosis},
  author={Guo, Yuanbo and Jia, Zhenge and Hu, Jingtong and Shi, Yiyu},
  booktitle={International Conference on Medical Image Computing and Computer-Assisted Intervention},
  pages={329--338},
  year={2024},
  organization={Springer}
}

@article{hardt2016equality,
  title={Equality of opportunity in supervised learning},
  author={Hardt, Moritz and Price, Eric and Srebro, Nati},
  journal={Advances in neural information processing systems},
  volume={29},
  year={2016}
}

@inproceedings{kamiran2012data,
  title={Data preprocessing techniques for classification without discrimination},
  author={Kamiran, Faisal and Calders, Toon},
  journal={Knowledge and information systems},
  volume={33},
  number={1},
  pages={1--33},
  year={2012},
  publisher={Springer}
}

@misc{asuncion2007uci,
  title={UCI machine learning repository},
  author={Asuncion, Arthur and Newman, David and others},
  year={2007},
  publisher={Irvine, CA, USA}
}

@article{flores2016false,
  title={False positives, false negatives, and false analyses: A rejoinder to machine bias: There's software used across the country to predict future criminals. and it's biased against blacks},
  author={Flores, Anthony W and Bechtel, Kristin and Lowenkamp, Christopher T},
  journal={Fed. Probation},
  volume={80},
  pages={38},
  year={2016},
  publisher={HeinOnline}
}

@inproceedings{puyol2021fairness,
  title={Fairness in cardiac MR image analysis: an investigation of bias due to data imbalance in deep learning based segmentation},
  author={Puyol-Ant{\'o}n, Esther and Ruijsink, Bram and Piechnik, Stefan K. and Neubauer, Stefan and Petersen, Steffen E. and Razavi, Reza and King, Andrew P.},
  booktitle={International Conference on Medical Image Computing and Computer-Assisted Intervention},
  pages={413--423},
  year={2021},
  organization={Springer}
}

@inproceedings{zhang2022improving,
  title={Improving the fairness of chest x-ray classifiers},
  author={Zhang, Haoran and Dullerud, Natalie and Roth, Karsten and Oakden-Rayner, Lauren and Pfohl, Stephen and Ghassemi, Marzyeh},
  booktitle={Conference on health, inference, and learning},
  pages={204--233},
  year={2022},
  organization={PMLR}
}

@article{zong2022medfair,
  title={MEDFAIR: benchmarking fairness for medical imaging},
  author={Zong, Yongshuo and Yang, Yongxin and Hospedales, Timothy},
  journal={arXiv preprint arXiv:2210.01725},
  year={2022}
}

@inproceedings{dwork2018decoupled,
  title={Decoupled classifiers for group-fair and efficient machine learning},
  author={Dwork, Cynthia and Immorlica, Nicole and Kalai, Adam Tauman and Leiserson, Max},
  booktitle={Conference on fairness, accountability and transparency},
  pages={119--133},
  year={2018},
  organization={PMLR}
}

@inproceedings{evgeniou2004regularized,
  title={Regularized multi--task learning},
  author={Evgeniou, Theodoros and Pontil, Massimiliano},
  booktitle={Proceedings of the tenth ACM SIGKDD international conference on Knowledge discovery and data mining},
  pages={109--117},
  year={2004}
}

@inproceedings{luo2022adapt,
  title={Adapt to adaptation: Learning personalization for cross-silo federated learning},
  author={Luo, Jun and Wu, Shandong},
  booktitle={IJCAI: proceedings of the conference},
  volume={2022},
  pages={2166},
  year={2022}
}

@article{tan2022towards,
  title={Towards personalized federated learning},
  author={Tan, Alysa Ziying and Yu, Han and Cui, Lizhen and Yang, Qiang},
  journal={IEEE transactions on neural networks and learning systems},
  volume={34},
  number={12},
  pages={9587--9603},
  year={2022},
  publisher={IEEE}
}

@inproceedings{dehdashtian2024utility,
  title={Utility-fairness trade-offs and how to find them},
  author={Dehdashtian, Sepehr and Sadeghi, Bashir and Boddeti, Vishnu Naresh},
  booktitle={Proceedings of the IEEE/CVF Conference on Computer Vision and Pattern Recognition},
  pages={12037--12046},
  year={2024}
}

@article{duan2025cost,
  title={The cost of local and global fairness in Federated Learning},
  author={Duan, Yuying and Xu, Gelei and Shi, Yiyu and Lemmon, Michael},
  journal={Proceedings of the 28th International Conference on Artificial Intelligence and Statistics (AISTATS)},
  year={2025}
}

@article{seyyed2021underdiagnosis,
  title={Underdiagnosis bias of artificial intelligence algorithms applied to chest radiographs in under-served patient populations},
  author={Seyyed-Kalantari, Laleh and Zhang, Haoran and McDermott, Matthew BA and Chen, Irene Y and Ghassemi, Marzyeh},
  journal={Nature medicine},
  volume={27},
  number={12},
  pages={2176--2182},
  year={2021},
  publisher={Nature Publishing Group US New York}
}

@article{daneshjou2022disparities,
  title={Disparities in dermatology AI performance on a diverse, curated clinical image set},
  author={Daneshjou, Roxana and Vodrahalli, Kailas and Novoa, Roberto A and Jenkins, Melissa and Liang, Weixin and Rotemberg, Veronica and Ko, Justin and Swetter, Susan M and Bailey, Elizabeth E and Gevaert, Olivier and others},
  journal={Science advances},
  volume={8},
  number={31},
  pages={eabq6147},
  year={2022},
  publisher={American Association for the Advancement of Science}
}

@article{xu2022algorithmic,
  title={Algorithmic fairness in computational medicine},
  author={Xu, Jie and Xiao, Yunyu and Wang, Wendy Hui and Ning, Yue and Shenkman, Elizabeth A and Bian, Jiang and Wang, Fei},
  journal={EBioMedicine},
  volume={84},
  year={2022},
  publisher={Elsevier}
}

@article{xu2024addressing,
  title={Addressing fairness issues in deep learning-based medical image analysis: a systematic review},
  author={Xu, Zikang and Li, Jun and Yao, Qingsong and Li, Han and Zhao, Mingyue and Zhou, S Kevin},
  journal={npj Digital Medicine},
  volume={7},
  number={1},
  pages={286},
  year={2024},
  publisher={Nature Publishing Group UK London}
}

@article{shazeer2017outrageously,
  title={Outrageously large neural networks: The sparsely-gated mixture-of-experts layer},
  author={Shazeer, Noam and Mirhoseini, Azalia and Maziarz, Krzysztof and Davis, Andy and Le, Quoc and Hinton, Geoffrey and Dean, Jeff},
  journal={arXiv preprint arXiv:1701.06538},
  year={2017}
}

@inproceedings{hu2018squeeze,
  title={Squeeze-and-excitation networks},
  author={Hu, Jie and Shen, Li and Sun, Gang},
  booktitle={Proceedings of the IEEE conference on computer vision and pattern recognition},
  pages={7132--7141},
  year={2018}
}

@article{de2017modulating,
  title={Modulating early visual processing by language},
  author={De Vries, Harm and Strub, Florian and Mary, J{\'e}r{\'e}mie and Larochelle, Hugo and Pietquin, Olivier and Courville, Aaron C},
  journal={Advances in neural information processing systems},
  volume={30},
  year={2017}
}

@inproceedings{guo2021parameter,
  title={Parameter-efficient transfer learning with diff pruning},
  author={Guo, Demi and Rush, Alexander M and Kim, Yoon},
  booktitle={Proceedings of the 59th Annual Meeting of the Association for Computational Linguistics and the 11th International Joint Conference on Natural Language Processing (Volume 1: Long Papers)},
  pages={4884--4896},
  year={2021}
}

@article{kim2022transfer,
  title={Transfer learning for medical image classification: a literature review},
  author={Kim, Hee E and Cosa-Linan, Alejandro and Santhanam, Nandhini and Jannesari, Mahboubeh and Maros, Mate E and Ganslandt, Thomas},
  journal={BMC medical imaging},
  volume={22},
  number={1},
  pages={69},
  year={2022},
  publisher={Springer}
}

@inproceedings{zhang2023robust,
  title={Robust mixture-of-expert training for convolutional neural networks},
  author={Zhang, Yihua and Cai, Ruisi and Chen, Tianlong and Zhang, Guanhua and Zhang, Huan and Chen, Pin-Yu and Chang, Shiyu and Wang, Zhangyang and Liu, Sijia},
  booktitle={Proceedings of the IEEE/CVF International Conference on Computer Vision},
  pages={90--101},
  year={2023}
}

@inproceedings{groh2021evaluating,
  title={Evaluating deep neural networks trained on clinical images in dermatology with the fitzpatrick 17k dataset},
  author={Groh, Matthew and Harris, Caleb and Soenksen, Luis and Lau, Felix and Han, Rachel and Kim, Aerin and Koochek, Arash and Badri, Omar},
  booktitle={Proceedings of the IEEE/CVF conference on computer vision and pattern recognition},
  pages={1820--1828},
  year={2021}
}

@misc{ODIR5Kdataset,
  title = {Ocular Disease Intelligent Recognition ODIR-5K},
  howpublished = {AcademicTorrents},
  year = {2019},
  url = {https://academictorrents.com/details/cf3b8d5ecdd4284eb9b3a80fcfe9b1d621548f72}
}

@article{pacheco2020pad,
  title={PAD-UFES-20: A skin lesion dataset composed of patient data and clinical images collected from smartphones},
  author={Pacheco, Andre GC and Lima, Gustavo R and Salomao, Amanda S and Krohling, Breno and Biral, Igor P and De Angelo, Gabriel G and Alves Jr, F{\'a}bio CR and Esgario, Jos{\'e} GM and Simora, Alana C and Castro, Pedro BC and others},
  journal={Data in brief},
  volume={32},
  pages={106221},
  year={2020},
  publisher={Elsevier}
}

@article{cubuk2018autoaugment,
  title={Autoaugment: Learning augmentation policies from data},
  author={Cubuk, Ekin D and Zoph, Barret and Mane, Dandelion and Vasudevan, Vijay and Le, Quoc V},
  journal={arXiv preprint arXiv:1805.09501},
  year={2018}
}

@inproceedings{xu2023fairadabn,
  title={Fairadabn: Mitigating unfairness with adaptive batch normalization and its application to dermatological disease classification},
  author={Xu, Zikang and Zhao, Shang and Quan, Quan and Yao, Qingsong and Zhou, S Kevin},
  booktitle={International Conference on Medical Image Computing and Computer-Assisted Intervention},
  pages={307--317},
  year={2023},
  organization={Springer}
}

@article{hemker2024healnet,
  title={Healnet: Multimodal fusion for heterogeneous biomedical data},
  author={Hemker, Konstantin and Simidjievski, Nikola and Jamnik, Mateja},
  journal={Advances in Neural Information Processing Systems},
  volume={37},
  pages={64479--64498},
  year={2024}
}

@inproceedings{xu2025fair,
  title={Fair Dermatological Disease Diagnosis Through Auto-weighted Federated Learning and Performance-Aware Personalization},
  author={Xu, Gelei and Wu, Yawen and Jia, Zhenge and Hu, Jingtong and Shi, Yiyu},
  booktitle={MICCAI Workshop on Fairness of AI in Medical Imaging},
  pages={167--176},
  year={2025},
  organization={Springer}
}

@inproceedings{he2016deep,
  title={Deep residual learning for image recognition},
  author={He, Kaiming and Zhang, Xiangyu and Ren, Shaoqing and Sun, Jian},
  booktitle={Proceedings of the IEEE conference on computer vision and pattern recognition},
  pages={770--778},
  year={2016}
}

@inproceedings{liu2021swin,
  title={Swin transformer: Hierarchical vision transformer using shifted windows},
  author={Liu, Ze and Lin, Yutong and Cao, Yue and Hu, Han and Wei, Yixuan and Zhang, Zheng and Lin, Stephen and Guo, Baining},
  booktitle={Proceedings of the IEEE/CVF international conference on computer vision},
  pages={10012--10022},
  year={2021}
}
